\documentclass[journal]{IEEEtran}

\usepackage{cite}

\ifCLASSINFOpdf
  \usepackage[pdftex]{graphicx}
\else
\fi
\usepackage{amsmath}
\usepackage{algorithmic}

\usepackage{array}
\usepackage{url}

\usepackage{amsfonts}
\usepackage{multirow}

\begin{document}
%
\title{IAN: The Individual Aggregation Network for Person Search }

%
%
%

\author{Jimin~XIAO,~\IEEEmembership{Member,~IEEE,}
        Yanchun~XIE,
        Tammam~TILLO,~\IEEEmembership{Senior Member,~IEEE,}
        Kaizhu~HUANG,~\IEEEmembership{Senior Member,~IEEE,}
        Yunchao~WEI,~\IEEEmembership{Member,~IEEE,}
        Jiashi~FENG

\thanks{Jimin~XIAO, Yanchun~XIE, Tammam~TILLO, and Kaizhu~HUANG are with the Department of Electrical and Electronic Engineering,
Xi'an Jiaotong-Liverpool University, Suzhou, P.R. China, e-mail: (jimin.xiao@xjtlu.edu.cn, yanchun.xie@xjtlu.edu.cn, tammam.tillo@xjtlu.edu.cn,kaizhu.huang@xjtlu.edu.cn).}
\thanks{Yunchao~WEI and Jiashi Feng are with the Department of Electrical and
Computer Engineering, National University of Singapore, Singapore, e-mail: wychao1987@gmail.com, elefjia@nus.edu.sg.}
}

\maketitle

\begin{abstract}

Person search in real-world scenarios is a
new challenging computer version task with many meaningful applications.
The challenge of this task mainly comes from:
(1) unavailable bounding boxes for pedestrians and the model needs to search for the person over the whole gallery images; (2) huge variance of
visual appearance of a particular person owing to varying
poses, lighting conditions, and occlusions. To address these two
critical issues in modern person search applications, we
propose a novel Individual Aggregation Network (IAN) that can
accurately localize persons by learning to minimize intra-person feature variations. IAN is built upon the state-of-the-art object detection framework,
i.e., faster R-CNN \cite{ren2015faster}, so that high-quality region proposals for
pedestrians can be produced in an online manner.
In addition, to relieve the negative effect caused by
varying visual appearances of the same individual, IAN introduces a novel center
loss that can increase the intra-class compactness of feature
representations. The engaged center loss encourages persons with the same identity to have similar feature characteristics.
Extensive experimental results on two benchmarks, i.e., CUHK-SYSU and PRW, well
demonstrate the superiority of the proposed model. In particular, IAN
achieves $77.23\%$ mAP and $80.45\%$ top-1 accuracy on CUHK-SYSU,
which outperform the state-of-the-art by $1.7\%$
and $1.85\%$, respectively.

\end{abstract}

\begin{IEEEkeywords}
person search, re-identification, pedestrian detection, softmax loss, center loss, dropout
\end{IEEEkeywords}

%
\IEEEpeerreviewmaketitle

\section{Introduction}

Person re-identification is to re-identify the same person across different cameras,
and it has attracted increasingly more interest in recent years ~\cite{bedagkar2014survey, gong2014person}.
The emergence of this task is mainly stimulated by (1) increasing demand of public security and (2) widespreading surveillance camera networks among public places, such as airports, universities, shopping malls, etc.
The obtained images from surveillance cameras are usually with some characteristics, e.g., low-quality, variable, and contain motion blur.
Traditional biometrics, such as face \cite{chatzis1999multimodal, dornaika2013exponential}, iris \cite{da2012dynamic}
and fingerprint \cite{cappelli2012fast}, are generally not available. Thus, many
person re-identification applications exploit the reliable body appearance.

\begin{figure*}
\begin{center}
   \includegraphics[width=  0.90 \linewidth, bb= 120 100 820 480, clip]{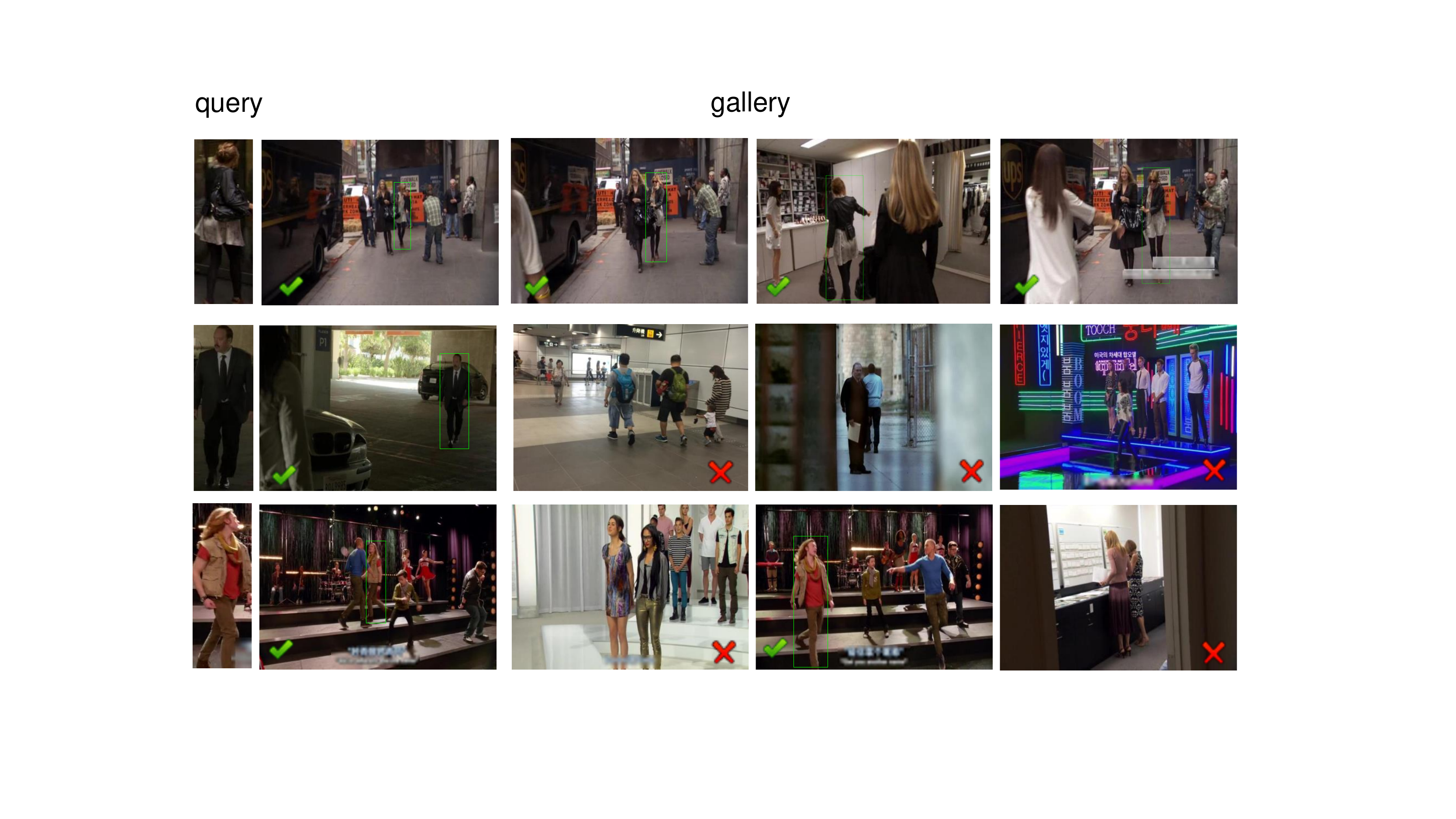}
\end{center}
   \caption{Person search from whole images without cropping out persons. The left column
   is probe/query image, other columns are gallery images without cropped pedestrians. The green bounding boxes are searching results.
   To find the right person in the gallery images, we need to detect all the persons within the image,
   and compare the detected persons with the probe image.
    }
\label{fig:demo1}
\end{figure*}

Technically, a person re-identification system for video surveillance consists of three components,
including person detection, person tracking, and person retrieval. While the first two components are
independent computer vision tasks, most person re-identification studies focus on the third component.
Numerous re-identification algorithms as well as datasets \cite{yi2014deep, li2014deepreid, zheng2011person, zheng2013reidentification, zheng2015scalable,tao2015person} have been proposed during the past decades and the performance on these benchmarks have been improved substantially.
All these algorithms focus on the third component of the pipeline, assuming the person/pedestrian detection are already available.
In other words, a query person is matched with cropped pedestrians in the gallery instead of searching for the target person from whole images.
In reality, perfect pedestrian bounding boxes are unavailable in surveillance scenarios. In addition, existing
pedestrian detectors unavoidably produce false alarms, misdetections, and misalignments.
All these factors compromise the re-identification performance. Therefore, current re-identification algorithms
cannot be directly applied to real surveillance systems,
where we need to search a person from whole images, as shown in Fig.~\ref{fig:demo1}.

While majority of person re-identification works engage boxes manually annotated or produced by a fixed detector in their applications,
it is necessary to study the impact of pedestrian detectors on re-identification accuracy. Specifically, \cite{xu2014person, zheng2016person}
showed that considering detection and re-identification jointly leads to higher person search accuracy than optimizing them separately.
To the best of our knowledge, end-to-end deep learning for person search (E2E-PS) \cite{xiao2016end}
is the first work to introduce an end-to-end deep learning
framework to jointly handle the challenges from both detection and re-identification.
Thereby, the detector and re-identification parts can interact with each other so as to reduce the influence of detection
misalignments.

\begin{figure}
\begin{center}
   \includegraphics[width=  \linewidth, bb= 50 150 800 500, clip]{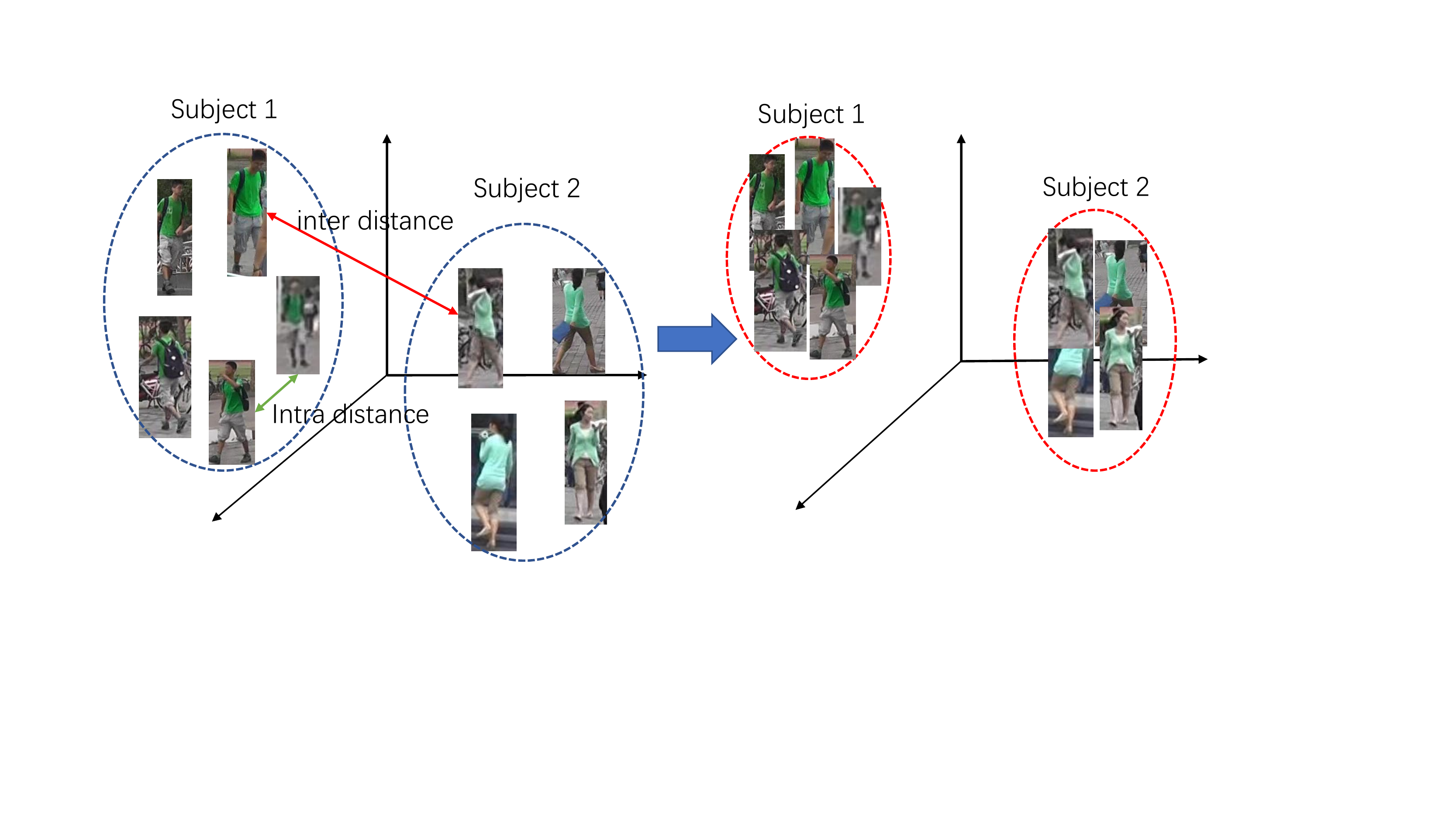}
\end{center}
   \caption{The objective of center loss is to reduce the intra-class distance by pulling the sample features towards each class center.
    Left side: feature distance without center loss; right side: feature distance using center loss.
    }
\label{fig:demo2}
\end{figure}

In E2E-PS \cite{xiao2016end}, the re-identification feature learning exploits a modified softmax loss. Early works show that such kind of identification task could greatly benefit the feature learning \cite{sun2014deep}. Meanwhile, it is found that the identification task increases the
inter-personal variations by drawing features extracted from different identities apart, while the verification task reduces the intra-personal
variations by pulling features extracted from the same identity together \cite{sun2014deep2}. Inspired by this, softmax loss and contrastive loss are jointly used for feature learning, leading to better performance than the sole softmax loss \cite{sun2014deep2}.
But we cannot directly introduce such verification tasks into the person search
faster R-CNN framework \cite{ren2015faster} used in E2E-PS \cite{xiao2016end}, since the pedestrians appearing in each image are random,
sparse, and unbalanced. It is difficult to organize equivalent amount of positive and
negative pedestrian pairs within the faster R-CNN framework.


To address this critical issue, we propose a novel Individual Aggregation Network (IAN) that can not
only accurately localize pedestrians but also minimize feature
representations of intra-person variations. In particular, IAN is built upon the state-of-the-art object detection framework,
i.e., Faster R-CNN \cite{ren2015faster}, so that high-quality region proposals for
pedestrians can be produced in an online manner for person search. In addition, to relieve the negative effect caused by
various visual appearances of the same individual, a novel center
loss \cite{wen2016discriminative} that can increase the intra-class compactness of feature representations is introduced.
The center loss encourages learned pedestrian representations from the same class to share similar feature characteristics.
The IAN model can be embedded in any CNN-based person search framework for improving the performance.

In particular, center loss is able to increase intra-class feature compactness without requiring to
aggregate positive and negative verification samples.
Center loss tracks the feature centers of all classes, and these feature centers are constantly
updated based on the recent observed class samples. Meanwhile, it manages to pull the sample features
towards each class center that this sample belongs to. This process is demonstrated in Fig.~\ref{fig:demo2}.
During the model development, we found that neural networks with dropout \cite{srivastava2014dropout} are not compatible with center loss \cite{wen2016discriminative}. We study this phenomenon in both analytic and experimental ways.
We believe this finding could be useful guidance for neural network framework design in the community, which is one of our contribution.

Finally, it is encouraging to see that our proposed IAN achieves
$77.23\%$ mAP and $80.45\%$ top-1 accuracy on the CUHK-SYSU person search dataset \cite{xiao2016end},
which is the new state-of-the-art for this dataset. Meanwhile, we also obtain state-of-the-art performance on the PRW dataset \cite{zheng2016person}.

The remainder of this paper is organized as follows. Related work is presented in Section \ref{sec:related_work}. The proposed person search method is described in Section \ref{sec:method}, with implementation details described in Section \ref{sec:implementation}. We present and discuss the experimental results in Section \ref{sec:experiment}. Finally, conclusions are draw in Section \ref{sec:conclusion}.

\section{Related Works}
\label{sec:related_work}

\noindent\textbf{Person re-identification}
CNN-based deep learning models have attracted a lot of attentions and
been successfully applied in person re-identification since two pioneer works \cite{yi2014deep,li2014deepreid}.
Generally, two categories of CNN models are commonly employed
in this community.  One cateorgy is the
Siamese model using image pairs \cite{yi2014deep,ahmed2015improved} or triplets \cite{ding2015deep} as input.
The other category is the classification model
as used in image classification and object detection.
Most re-identification datasets provide only two images for each pedestrian
such as VIPeR \cite{gray2008viewpoint}, CUKH01 \cite{li2012human} CUHK03 \cite{li2014deepreid},
therefore, currently most CNN-based re-identifications schemes use the Siamese model.
In \cite{yi2014deep}, an input image is partitioned into
three overlapping horizontal parts, and the parts go through
two convolutional layers plus one fully connected layer which
fuses them and outputs a vector to represent this image,
and lastly, two vectors are connected by a cosine layer.
Ahmed et al. \cite{ahmed2015improved} improved the Siamese model by computing the cross-input
neighborhood difference features, which compares the
features from one input image to features in neighboring
locations of the other image. In \cite{varior2016siamese}, Varior et al. incorporate long short-term memory (LSTM)
modules into a Siamese network so that the spatial connections can be memorized
to enhance the discriminative ability of the deep features.
Similarly, Liu et al. \cite{liu2016end} propose to integrate a soft attention based model in
a siamese network to adaptively focus on the important local
parts of the input image pair.

One disadvantage of the Siamese model is that it cannot take full advantage
of the re-identification annotations. The Siamese model only considers pairwise labels
(similar or not similar), which is a weak label.
Another potentially effective strategy is to use a classification/identification mode,
which makes full use of the re-identification labels.
On large datasets, such as PRW and MARS, the
classification model achieves good performance without
careful pairwise or triplet selection \cite{zheng2016mars, zheng2016person}.
In this paper, our identification method is also built based on a classification/identification mode.

\vspace{3ex}
\noindent\textbf{Pedestrian Detection}
In the past year, a lot of efforts have been made to improve the performance of pedestrian detection \cite{dollar2009integral, marin2014occlusion}.
The Integrate Channel Features (ICF) detector  \cite{dollar2009integral}
is among the most popular pedestrian detectors without using deep learning features.
Following its success,
many variants \cite{dollar2014fast,nam2014local} were proposed
with significant improvement.
Recent years, CNN-based pedestrian detectors have
also been developed. Various factors, including CNN model
structures, training data, and different training strategies are
studied empirically in \cite{hosang2015taking}. In \cite{ zhang2016faster},
faster R-CNN is studied for pedestrian detection.

\section{Individual Aggregation Network}
\label{sec:method}

\begin{figure*}
\begin{center}
   \includegraphics[width= 0.9 \linewidth, bb= 40 90 800 420, clip]{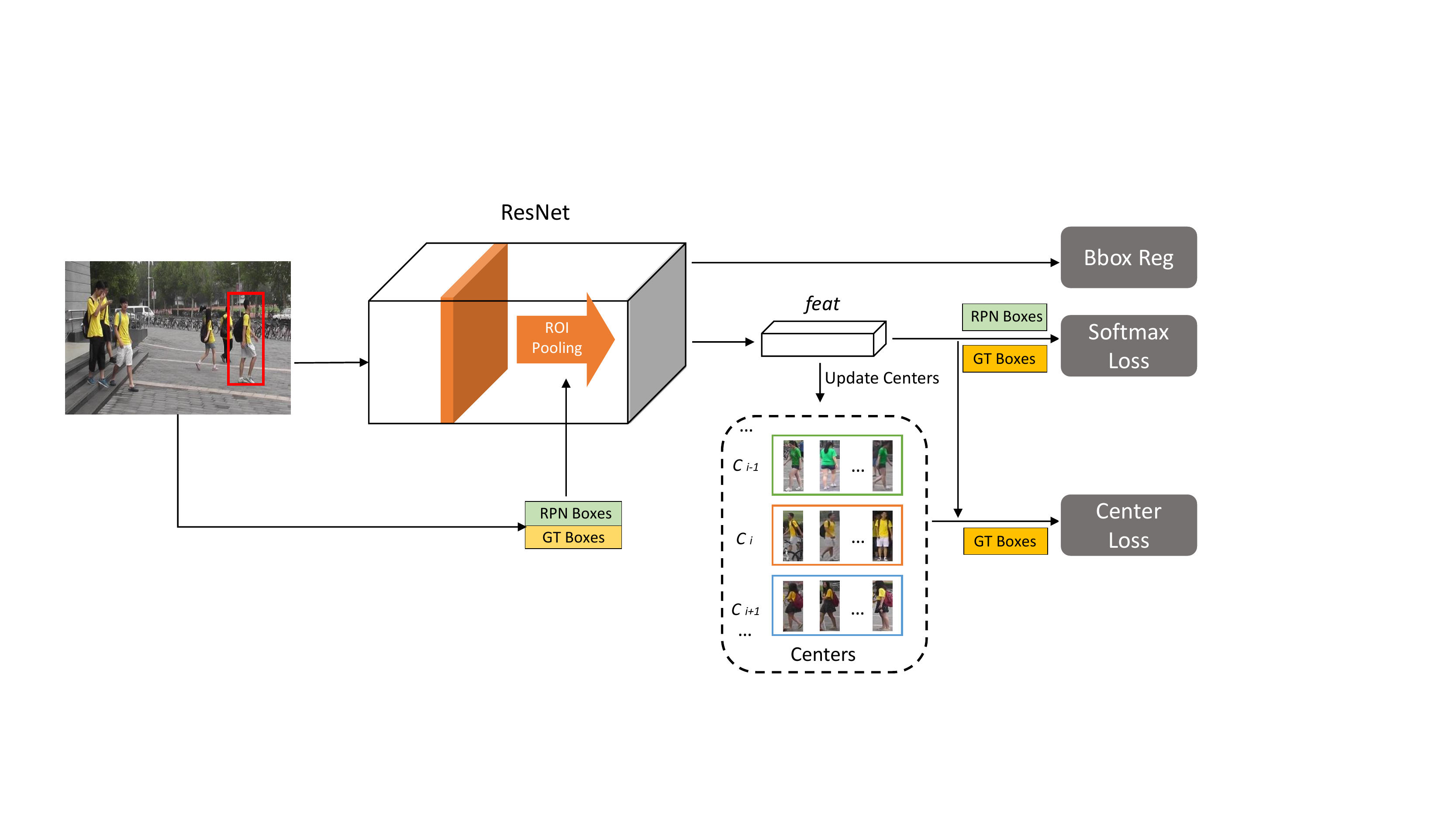}
\end{center}
   \caption{Overview of our IAN network training framework. Images containing pedestrians are input into the network.
   First part of residual network, i.e., layers Conv1-Res4b of ResNet-101 \cite{He2015}, is used to get image features.
   Pedestrian bounding boxes generated by region proposal network (RPN boxes),
   together with the ground truth pedestrian bounding boxes (GT boxes), are used
   for ROI pooling to generate feature vector for each box. Second part of residual network,
   i.e., layers Res5a-Res5c of ResNet-101 \cite{He2015}, uses the ROI pooling features as input. Two fully connected layers
   are utilized separately, one to produce the final feature vector \emph{feat} to compute feature distance,
   and another to produce bounding box locations. Feature vectors of all candidates boxes (RPN + GT boxes) go
   into the random sampling softmax loss layer; while only feature vectors of ground truth pedestrian boxes (GT boxes) go
   into the center loss layer.
    }
\label{fig:overall}
\end{figure*}

In practical person search applications, pedestrian bounding boxes
are unavailable and the target
person needs to be found from the whole images.
Targeting this problem, IAN is built upon the state-of-the-art object detection framework, i.e., faster R-CNN \cite{ren2015faster},
so that reasonable region proposals for
pedestrians can be produced in an online manner for person search.
The proposed IAN framework is shown in Fig.~\ref{fig:overall}, and
it is elaborated as follows.

\begin{enumerate}
\item
In the training phase, arbitrary size images with ground truth pedestrian bounding boxes and identifications are input into
the first part residual network \cite{He2015}. The residual network is divided into
$2$ parts, i.e, for ResNet-101 network, layers Conv1-Res4b forms the first part network, while layers Res5a-Res5c
are the second part.

\item
The region proposal network (RPN) \cite{ren2015faster}, is built on top of the feature maps generated
with the first part network to predict pedestrian bounding boxes.
The RPN is trained with ground truth pedestrian bounding boxes, using two loss layers, i.e., anchor classification and anchor regression.
Besides the candidate boxes generated by the region proposal network (RPN boxes), the ground truth (GT)
pedestrian bounding boxes are also used together at the network training stage. At the test stage, only RPN boxes are available.

\item
All the candidate boxes (RPN+GT boxes at training stage, RPN boxes at test stage)
are used for ROI pooling to generate feature vector for each candidate box.
These features are again convolved with the second part residual networks, i.e., layers Res5a-Res5c of ResNet-101.

\item
Two sibling fully connected layers are utilized separately, one to produce the final feature vector \emph{feat} to compute feature distance,
and the other to produce bounding box locations. At training stage, feature vectors of all candidates boxes (RPN+GT boxes) are fed
into the softmax loss layer, while only feature vectors of ground truth pedestrian boxes (GT boxes) are fed
into the center loss layer. The softmax variant random sampling softmax (RSS) \cite{xiao2016end} is used for training.

\end{enumerate}

Overall, compared with previous person search method E2E-PS \cite{xiao2016end}, the proposed IAN
generates more discriminative feature representations.
In IAN, using softmax loss together with center loss \cite{wen2016discriminative} within
the faster R-CNN framwork leads to better feature representations than solely using softmax loss in \cite{xiao2016end}.
Meanwhile, the VGGNet \cite{Simonyan14c} used in E2E-PS \cite{xiao2016end} contains dropout layers which are
intrinsically not compatible with the center loss.
In our IAN, we use the state-of-the-art residual network \cite{He2015}.
In addition to solving the compatibility issue with center loss,
replacing VGGNet with residual network also offers better discrimination
power with lower computational cost.

\subsection{Softmax  + Center Loss}

Both compact intra-class variations and separable inter-class differences are essential for discriminative features.
However, the softmax loss only encourages the separability of features.
Contrastive loss \cite{Hadsell06dimensionalityreduction, sun2014deep2} and
triplet loss \cite{Ding2015}, that respectively construct loss functions for image pairs and triplets,
are possible solutions to encourage intra-class variation compactness.
For contrastive loss, equivalent amount of positive and negative image pairs are required,
whereas for triplet loss, two images among each triplet should belong to the same class/identification with
one belonging to different class/identification.
However, for the faster R-CNN based person search framework, it is a non-trivial task to form such
image pairs and triplets within the input mini-batch.
The pedestrians within each image belongs to different identifications.
Meanwhile, the pedestrians appearing in each image are random, sparse and unbalanced.
Within the mini-batch of faster R-CNN, it is difficult to form a balanced number of positive pedestrian pairs as negative pairs.

On the other hand, employing center loss \cite{wen2016discriminative} is able to avoid the need of aggregating positive and negative pairs.
In the proposed IAN network, center loss is applied together with softmax loss to generate feature representations.
The center loss function is defined as follows.
\begin{equation}
\mathcal{L}_c = \frac{1}{2} \sum_{i = 1}^m \left\|x_i- c_{y_i} \right\|^2
\label{eqn:center_loss}
\end{equation}
where $x_i \in \mathbb{R}^d$ is the feature vector of pedestrian box $i$, which belongs to class $y_i$,
and $c_{y_i} \in \mathbb{R}^d $ denotes the $y_i$-th class center of features.
The softmax loss forces the features of different classes staying apart.
The center loss pulls the features of the same class closer to their centers.
Hence the feature discriminative power is highly enhanced.
With the center loss, the overall network loss function is defined as:
\begin{equation}
\mathcal{L} = \mathcal{L}_4 + \lambda \mathcal{L}_c
\label{eqn:five_loss}
\end{equation}
where $\mathcal{L}_4$ is the summation of $4$ loss functions in faster R-CNN, which includes the
softmax loss for perosn identification classification, and $\lambda$ is the weight of the center loss.

Ideally,  $c_{y_i}$ should be constantly updated as the network parameters are being updated.
In other words, we need to take the entire training set into account and average the features of every class in each iteration,
which is inefficient and impractical.
In fact, we learn the  feature center of each class. In the training process,
we simultaneously update the center and minimize the distances between the features
and their corresponding class centers.

The center $c_{y_i}$ is updated based on each mini-batch. In each iteration, the centers are computed by
averaging the features of the corresponding classes.
Meanwhile, to avoid large perturbations caused by few
mislabelled samples, we use a scalar $\alpha \in [0, 1] $ to control the learning rate of the centers.
The gradients of $\mathcal{L}_c$ with respect to $x_i$ and the updating equation of $c_{y_i}$ are computed
as:
\begin{equation}
\frac{\partial \mathcal{L}_c}{\partial x_i} = x_i - c_{y_i}
\label{eqn:feature_update}
\end{equation}
\begin{equation}
\Delta c_j = \frac{\sum_{i=1}^m \delta(y_i = j)\cdot (c_j-x_i) }{ 1+ \sum_{i=1}^m \delta(y_i = j)}
\label{eqn:center_update}
\end{equation}
where $\delta(\emph{condition}) = 1$ if the $\emph{condition}$ is satisfied, and otherwise
$\delta(\emph{condition}) = 0$.

\subsection{Why to Avoid Dropout?}
\label{sec:avoid_dropout}

In our study, we notice that neural networks with dropout are not compatible with the center loss.
For example, when the proposed IAN is deployed on VGGNet with $3$ dropout layers, its person search mAP
performance on the CUHK-SYSU person search dataset \cite{xiao2016end} is about $10\%$ lower than the results obtained by removing all the dropout layers.

Dropout is a technique for addressing overfitting problems \cite{srivastava2014dropout}. The key idea of dropout is to
randomly drop units, along with their connections, from the neural network during training.
Since the dropout randomly drops units, it creates uncertainty for the features. In other words,
when image features are extracted using the same network with dropout, the obtained features for the same image might be
quite different in different network forward computation instances. This is contradicting with center loss, which punishes intra-class
variations.

The dropout is usually deployed after the fully connect layer, as in VGGNet.
Let $z^{(l)}$ denote the vector of inputs into layer $l$, and $y^{(l)}$ denote the vector of outputs
from layer $l$. $W^{(l)}$ and $b^{(l)}$ are the weights and biases at layer $l$, respectively.
The feed-forward operation of a standard neural network can be described as
\begin{equation}
 z_i^{(l+1)} =  w_i^{(l+1)} y^{(l)} + b_i^{(l+1)}
\label{eqn:forward1}
\end{equation}
\begin{equation}
 y_i^{(l+1)} = f(z_i^{(l+1)})
\label{eqn:forward2}
\end{equation}
where $f$ is any activation function, for example sigmoid or ReLu function.
With dropout, the feed-forward operation becomes
\begin{equation}
 r_j^{(l)} \sim \emph{Bernoulli}(p)
\label{eqn:forward_d1}
\end{equation}
\begin{equation}
{\tilde{y}}^{(l)} = r^{(l)}*y^{(l)}
\label{eqn:forward_d2}
\end{equation}
\begin{equation}
 z_i^{(l+1)} =  w_i^{(l+1)} {\tilde{y}}^{(l)} + b_i^{(l+1)}
\label{eqn:forward_d3}
\end{equation}
\begin{equation}
 y_i^{(l+1)} = f(z_i^{(l+1)})
\label{eqn:forward_d4}
\end{equation}
Here $*$ denotes an element-wise product. For any layer $l$, $r^{(l)}$
is a vector of independent Bernoulli random variables each of which has probability $p$ of being $1$.

To illustrate that dropout is not compatible with the center loss, let us take one example.
Assume input image samples $I_i$ and $I_j$ are the same and belong to the same pedestrian/class.
We assume layer $l+1$ is a fully connection layer with
dropout, the output of layer $l+1$ is input into the center loss layer. Since
image samples $I_i$ and $I_j$ are the same, we could have
$y^{(l)}(I_i)  = y^{(l)}(I_j)$.
The target of the center loss is to have similar features for
the same class, i.e.,  $y^{(l+1)}(I_i)  = y^{(l+1)}(I_j)$.
Considering (\ref{eqn:forward_d2})(\ref{eqn:forward_d4}),
it is equivalent as $ r^{(l)}(I_i)*y^{(l)}(I_i) = r^{(l)}(I_j)*y^{(l)}(I_j)$.
Here $r^{(l)}(I_i)$ and $r^{(l)}(I_j)$ are vectors of independent \emph{Bernoulli} random variables,
leading to $r^{(l)}(I_i) \neq r^{(l)}(I_j)$.
Therefore, to have $y^{(l+1)}(I_i)  = y^{(l+1)}(I_j)$, the only solution is
$y^{(l)}(I_i) = y^{(l)}(I_j) = \vec{0}$.
However, zero feature cannot properly represent the image samples.
From the above simple example, we could conclude that dropout is not compatible with center loss,
which is consistent with our experimental verification.

\section{Implementation Details}
\label{sec:implementation}

\subsection{Training Phase}
\label{sec:training_phase}

During the network training phase, the network is trained to detect pedestrians and
produce discriminative features for re-identification.
In our network, $5$ loss functions are used.
The smoothed-L1 loss \cite{girshick2015fast} is used for the two bounding box regression layers.
A softmax loss is used for the pedestrian proposal module, which classifies pedestrian and non-pedestrian.
For the re-identification feature extraction part, we deploy both random sampling softmax \cite{xiao2016end}
and center loss \cite{wen2016discriminative}. Here it is important to note that only features of
ground truth pedestrian boxes are input into the center loss layer. This helps to avoid sample noise.
The overall loss is the sum of all five loss functions, and its gradient w.r.t. the network parameters
is computed through back propagation.

To speed up the network convergence, the training process includes three steps:
\begin{enumerate}
\item
We crop ground truth bounding boxes for each training person and randomly sample the same number of background boxes.
Then we shuffle the boxes, resize them to $224\times 224$, and fine-tune the residual network
model (ResNet-101 and ResNet-50) to classify the candidate boxes.
The output feature size of ROI-pooling layer in Fig.~\ref{fig:overall} is $7\times7$. To insure the same feature size,
we add one $2\times2$ pooling layer to the residual network.

\item
We fine-tune the model resulting from the above step. Unlike the previous step, the whole images with GT pedestrian bounding boxes and identification annotations are used for the fine-turning process. $4$ loss layers excluding the center loss is used in this
fine-tuning process.

\item
We fine-tune the model obtained in Step $2$ with all $5$ loss layers including the center loss.
The input images and label annotations are the same as those in Step $2$.

\end{enumerate}

\subsection{Test Phase}

The test phase is similar to that in \cite{xiao2016end}.
For each gallery image, we get the features (\emph{feat}) of all the candidate pedestrians by performing
the network forward computation once. Whereas for
the query image, we replace the pedestrian proposals with
the given bounding box, and then do the forward computation
to get its feature vector (\emph{feat}). Finally, we compute
the pairwise Euclidean distances between the query features
and those of the gallery candidates.
The person similarity level is evaluated based on the Euclidean distances.

\section{Experiments}
\label{sec:experiment}

\noindent\textbf{Dataset and Evaluation Metrics} We use the benchmark datasets,
i.e., both the CUHK-SYSU person search dataset \cite{xiao2016end}
and PRW dataset \cite{zheng2016person} in our experiment. Both mean Averaged Precision (mAP) and top-1 matching rate
metrics are used. A candidate window is considered as positive if it overlaps with
the ground truth larger than $0.5$, which is the same as the setup in previous works \cite{xiao2016end,zheng2016person}.

CUHK-SYSU dataset is a large scale and scene-diversified person search dataset, which
contains 18,184 images, 8,432 persons, and 99,809 annotated bounding boxes.
Each query person appears in at least two images. Each image may contain more than one query
person and many background people. The dataset is partitioned into a training set and a test set.
The training set contains 11,206 images and 5,532 query persons. The test
set contains 6,978 images and 2,900 query persons.
The training and test sets have no overlap on images or query
persons. The identifications in CUHK-SYSU dataset  is in the range of $[-1, 5532]$, with
$-1$ being unknown persons, and 5,532 being background.
$-1$ boxes do not go into the random sampling softmax (RSS).
Neither $-1$ nor 5,532 goes into the center loss layer, because unknown persons and background are not unique as other identifications.

In the PRW dataset, a total of 11,816 frames are manually annotated to obtain 43,110 pedestrian
bounding boxes, among which 34,304 pedestrians are annotated with an identifications
ranging from $1$ to $932$ and the rest are assigned an identifications of $-2$.
The PRW dataset is divided into a training set with 5,704 frames and 482 identifications
and a test set with 6,112 frames and 450 identifications. Similar to that in CUHK-SYSU dataset,
unknown persons, and background does not go into the center loss layer.
$-1$ boxes do not go into the random sampling softmax (RSS).

Our ablation study is based on the CUHK-SYSU dataset, so as to provide more comprehensive
performance comparisons with state-of-the-art methods, such as  E2E-PS \cite{xiao2016end} and JDI-PS \cite{xiao2017end}.

\vspace{3ex}
\noindent\textbf{Training/Testing Settings}
We build our framework on two residual networks, i.e., ResNet-101 and ResNet-50 \cite{He2015}.
For ResNet-101, the pedestrian proposal
network is connected after layer res4b22, while for ResNet-50, it is connected after layer res4f.
In the following experiments, the default network is  ResNet-101 if not specified.
For training Step $1$ described in Section \ref{sec:training_phase}, the learning rate is $0.001$ with
$20$k iterations and batch size being $8$.  For training Step $2$, $120$k iterations are used.
The initial learning rate is $0.001$ and decreased by a factor of 10 after $100$k iterations.
For training Step $3$, the learning rate is $0.0001$ with $20$k iterations.
For both steps $2$ and $3$, the batch size is $2$ due to high memory cost.
The networks are trained on NVIDIA GeForce TITAN X GPU with 12GB memory.
Our implementation is based on the publicly available Caffe framework \cite{jia2014caffe}.

For testing the CUHK-SYSU dataset, in order to evaluate the influence of gallery size,
different gallery size is used, including $\{50, 100, 500, 1000, 2000, 4000\}$.
In the following experiments, we will report the performance based on the test protocol where
the gallery size is $100$ if not specified.
Each image contains $5.3$ background persons on average. If the gallery size is set
to $100$, a query person has to be distinguished from around
$530$ background persons and thousands of non-pedestrian
bounding boxes, which is challenging.
While for testing the PRW dataset, all 6,112 frames in the test set are used as gallery,
which is challenging.

\begin{figure}
\begin{center}
   \includegraphics[width= \linewidth, bb= 130 260 470 530, clip]{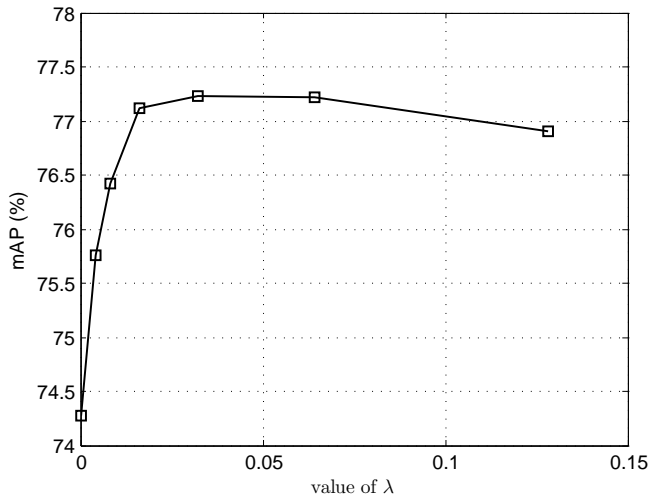}
\end{center}
   \caption{ The mAP accuracy of person search on CUHK-SYSU \cite{xiao2016end} validation set using different center loss weight $\lambda$.}
\label{fig:lambda}
\end{figure}

\subsection{Results on CUHK-SYSU dataset}
\label{sec:cuhk_sysu}

\noindent\textbf{Experiment on Parameter $\lambda$.} The hyper parameter $\lambda$
controls the weight of the center loss over the whole network loss function. It is essential to our model.
So we conduct one experiment to investigate the sensitiveness of of the proposed approach with respect to  $\lambda$.
We vary $\lambda$ from $0$ to $0.128$ to learn different models.
The training dataset is equally divided into $5$ equal folds, use $4$ of them for training, and $1$ for validation.
Cross-validation is deployed.
The person search accuracies of these models on CUHK-SYSU \cite{xiao2016end} validation set are
shown in Fig.~\ref{fig:lambda}. It is very clear that it is not a good choice simply without using the
center loss (in this case $\lambda=0$), leading to poor person search mAP performance.
Proper choice of the values, e.g., $\lambda = 0.032$, can improve the person search accuracy of the deeply
learned features. We also observe that the person search performance of our model
remains largely stable across a wide range $[0.016, 0.128]$.
Meanwhile, it is also observed that similar trend is obtained for the top-1 accuracy.
Thus, in the following experiments, we set the $\lambda$ value as $0.032$.

\vspace{3ex}
\noindent\textbf{Overall Person Search Performance.}
The results of IAN and benchmarks under two evaluation metrics are summarized in Table \ref{tab:performance}.
We compare our performance with end-to-end deep learning for person search (E2E-PS) method \cite{xiao2016end},
and joint detection and identification feature learning for person search (JDI-PS) method \cite{xiao2017end},
because of their superior performance.
As reported in \cite{xiao2017end}, JDI-PS method \cite{xiao2017end} attains much better performance
than separating pedestrian detection (\cite{dollar2014fast}, \cite{yang2015convolutional})
and re-identification (for examples, BoW \cite{koestinger2012large}+ Cosine similarity, LOMO+XQDA \cite{liao2015person}).

With ResNet-101, more than $7\%$ gain is obtained compared with \cite{xiao2016end} for both mAP and
top-1 accuracy. To demonstrate the importance of center loss in IAN, we also report the performance of
E2E-PS \cite{xiao2016end} when the VGGNet is replaced with  ResNet-101 and ResNet-50.
It is observed that $2\%-3\%$ gain for the two metrics is obtained only because of the center loss.
It is important to note that our performance is also better than JDI-PS \cite{xiao2017end} if both
deploy the ResNet-50.

\begin{table*}
\begin{center}
\caption{Comparisons between IAN with E2E-PS \cite{xiao2016end} and JDI-PS \cite{xiao2017end}.}
\label{tab:performance}
\begin{tabular}{|c|c|c|c|c|c|c|}
\hline
Method & E2E-PS \cite{xiao2016end} &  E2E-PS \cite{xiao2016end}   &  E2E-PS \cite{xiao2016end}  & JDI-PS \cite{xiao2017end}   & IAN  & IAN  \\
& (VGGNet) & (ResNet-50) & (ResNet-101) & (ResNet-50) & (ResNet-50) & (ResNet-101) \\
\hline\hline
mAP (\%) & $69.69$ & $73.13$ & $74.28$     &　$75.5$　&　$76.28$ &  \textbf{77.23} \\
top-1 (\%) & $72.97$ & $77.34$  & $78.17$  &　$78.7$　& $80.07$ & \textbf{80.45} \\
\hline
\end{tabular}
\end{center}
\end{table*}

\vspace{3ex}
\noindent\textbf{Input of Center Loss.}
In our proposed method, only features of ground truth pedestrian boxes are input into the center loss layer.
This scheme is verified by experimental results. To do this, we input all positive pedestrian boxes
(excluding background and unknown persons with id $-1$) into the center loss layer. Note that positive pedestrian boxes
refer to candidate boxes overlapping with ground truth pedestrian boxes higher than threshold, i.e., $0.5$.
The obtained results with such scheme is lower than that uses features of ground truth pedestrian boxes, as reported Table \ref{tab:groundtruth}.
This is because the objective of center loss is to increase intra-class feature compactness, but features of
different positive boxes of the same pedestrian are dissimilar as they cover different regions with various background information.

\begin{table}
\begin{center}
\caption{The person search performance if all positive pedestrian boxes
are input into the center loss layer (IAN with all boxes). }
\label{tab:groundtruth}
\begin{tabular}{|c|c|c|c|c|c|}
\hline
Method & IAN with all boxes & IAN  \\
\hline\hline
mAP (\%) & $74.70$  &  77.23 \\
top-1 (\%) & $77.72$  & 80.45 \\
\hline
\end{tabular}
\end{center}
\end{table}

\vspace{3ex}
\noindent\textbf{Center Loss with VGGNet.}
In Section \ref{sec:avoid_dropout}, analysis to avoid dropout is given. We also
study this phenomenon with experiments. The VGGNet model provided in \cite{xiao2016end},
where dropout layers are used, is fine-turned with center loss with loss weight $0.0032$.
The testing results with the fine-turned models are reported in Table \ref{tab:dropout}.
It is observed that by increasing the iteration number, the performance is decreased
constantly. With $40,000$ iterations, almost $9\%$ mAP is dropped compared with models without
center loss. The importance of replacing VGGNet with ResNet is demonstrated with this experiment.

\begin{table}
\begin{center}
\caption{Person search performance using VGGNet (dropout) and center loss together.}
\label{tab:dropout}
\begin{tabular}{|c|c|c|c|c|c|}
\hline
Iteration & 0 & $10,000$ & $20,000$ & $30,000$  & $40,000$    \\
\hline\hline
mAP (\%) & $69.69$ & $67.38$   & $64.12$ &  $62.55$ &  $60.73$ \\
top-1 (\%) & $72.97$ & $71.31$  & $69.03$  & $66.79$ &  $66.21$ \\
\hline
\end{tabular}
\end{center}
\end{table}

We remove all the dropout layers in VGGNet, and test E2E-PS\cite{xiao2016end} and
our IAN. The obtained results are
reported in Table \ref{tab:vgg_nodrop}. It is interesting to see that removing the $3$ dropout layers in
VGGNet leads to sightly better person search performance. Our IAN with center loss
leads to about $2\%$ performance gain compared with E2E-PS\cite{xiao2016end} for both mAP and top-1 accuracy,
if both remove the dropout layers. By comparing the results in Table \ref{tab:dropout} and \ref{tab:vgg_nodrop},
it is evident that dropout and center loss are not compatible. The experimental results
support our analysis in Section \ref{sec:avoid_dropout}.

\begin{table}
\begin{center}
\caption{Comparison between IAN and E2E-PS \cite{xiao2016end} for VGGNet with all dropout layers removed.}
\label{tab:vgg_nodrop}
\begin{tabular}{|c|c|c|c|c|c|c|}
\hline
Method & E2E-PS\cite{xiao2016end} &  E2E-PS\cite{xiao2016end} &  IAN  \\
& (VGGNet)  & (VGGNet no dropout) & (VGGNet) \\
\hline\hline
mAP (\%)   & $69.69$ &  $71.21$ &  \textbf{73.65} \\
top-1 (\%) & $72.97$ &  $74.48$ &  \textbf{76.14} \\
\hline
\end{tabular}
\end{center}
\end{table}

\begin{figure}
  \centering
    \includegraphics[width=  \linewidth, bb= 140 260 450 520, clip]{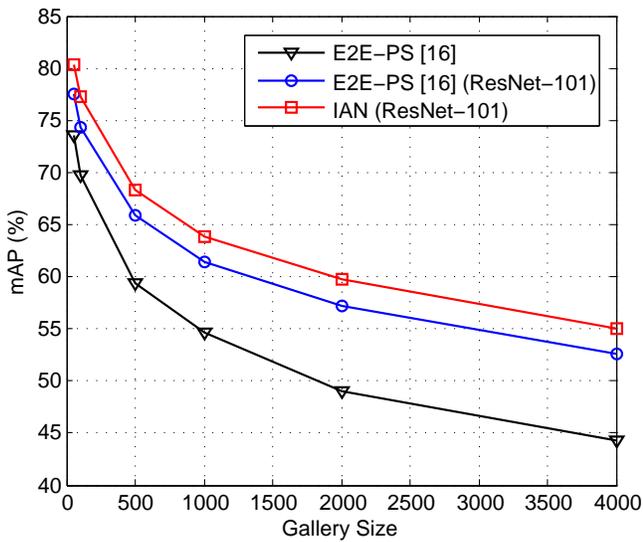}
    \centerline{\small (a)  }
    \includegraphics[width=  \linewidth, bb= 140 260 450 520, clip]{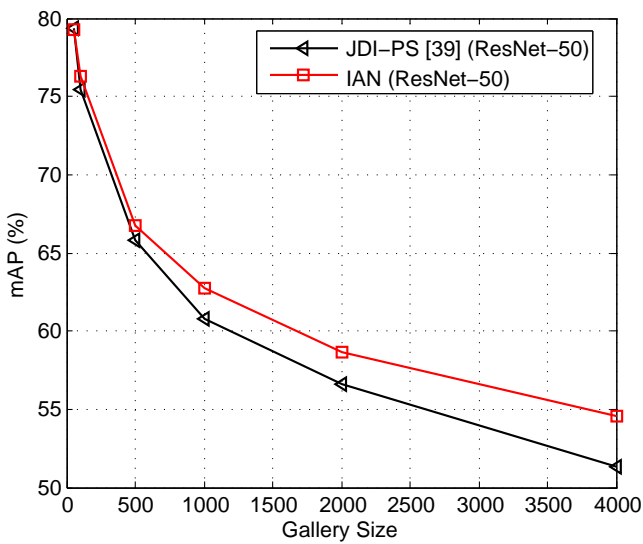}
   \centerline{\small (b)  }
   \caption{Person search performance comparison for various gallery size.
   (a) Comparing IAN with E2E-PS \cite{xiao2016end}; (b) Comparing IAN with JDI-PS \cite{xiao2017end}. }
\label{fig:gallery_size}
\end{figure}

\vspace{3ex}
\noindent\textbf{Effects of Gallery Size.} The task of person search is more challenging
when the gallery size increases. We vary the gallery size from $50$ to 4,000,
and test our approach, E2E-PS \cite{xiao2016end} with both VGGNet and ResNet-101, and JDI-PS \cite{xiao2017end}.
The obtained mAPs for various  gallery size are reported in Fig.~\ref{fig:gallery_size}.
As expected, the mAP decreases with the increase of gallery size.
Meanwhile, for various  gallery sizes, our approach outperforms E2E-PS \cite{xiao2016end}
with both VGGNet and ResNet-101 significantly. For large gallery size 4,000,
the mAP gain over E2E-PS \cite{xiao2016end} is more than $10\%$.
Meanwhile, it is also observed from Fig.~\ref{fig:gallery_size}.(b) that IAN outperforms
JDI-PS \cite{xiao2017end} with good gain for various gallery size. For large gallery size, i.e., 4,000,
the mAP gain is $3\%$. It is worth noticing that the comparison is fair because both use the ResNet-50 network.

\begin{figure*}
\begin{center}
   \includegraphics[width=  \linewidth, bb= 70 130 940 420, clip]{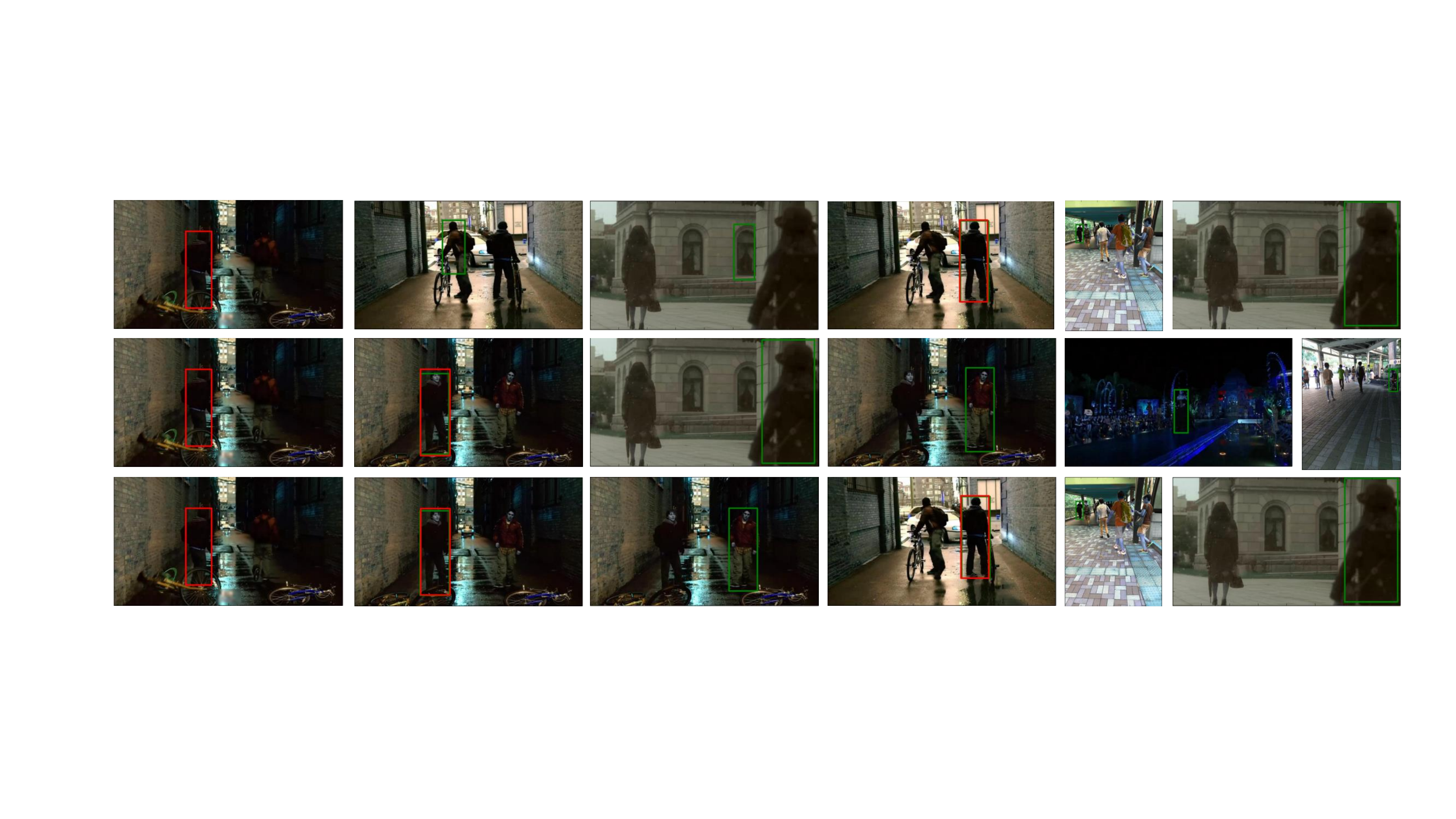} \\
   \includegraphics[width=  \linewidth, bb= 10 140 895 410, clip]{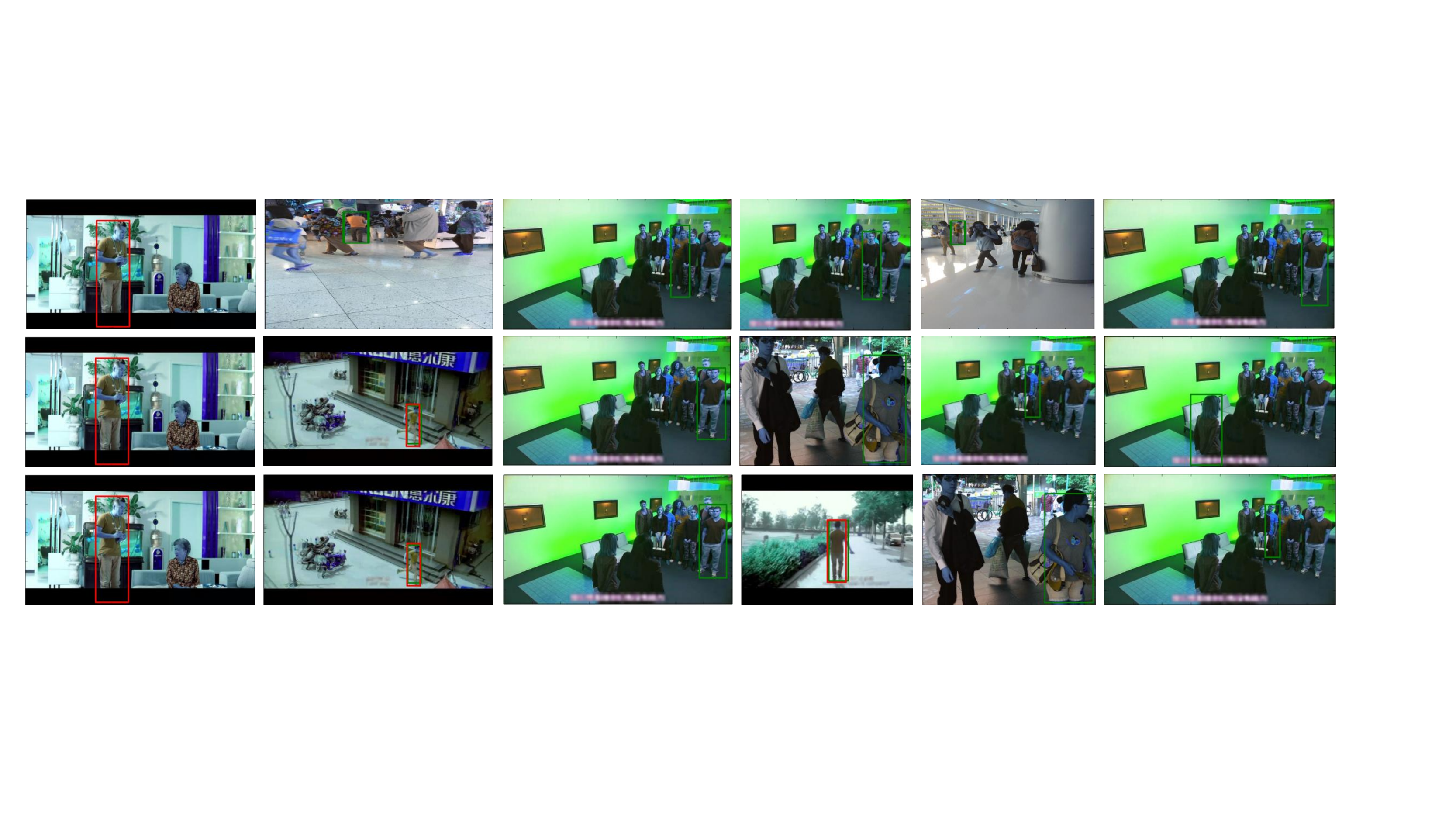} \\
   \includegraphics[width=  \linewidth, bb= 70 130 910 420, clip]{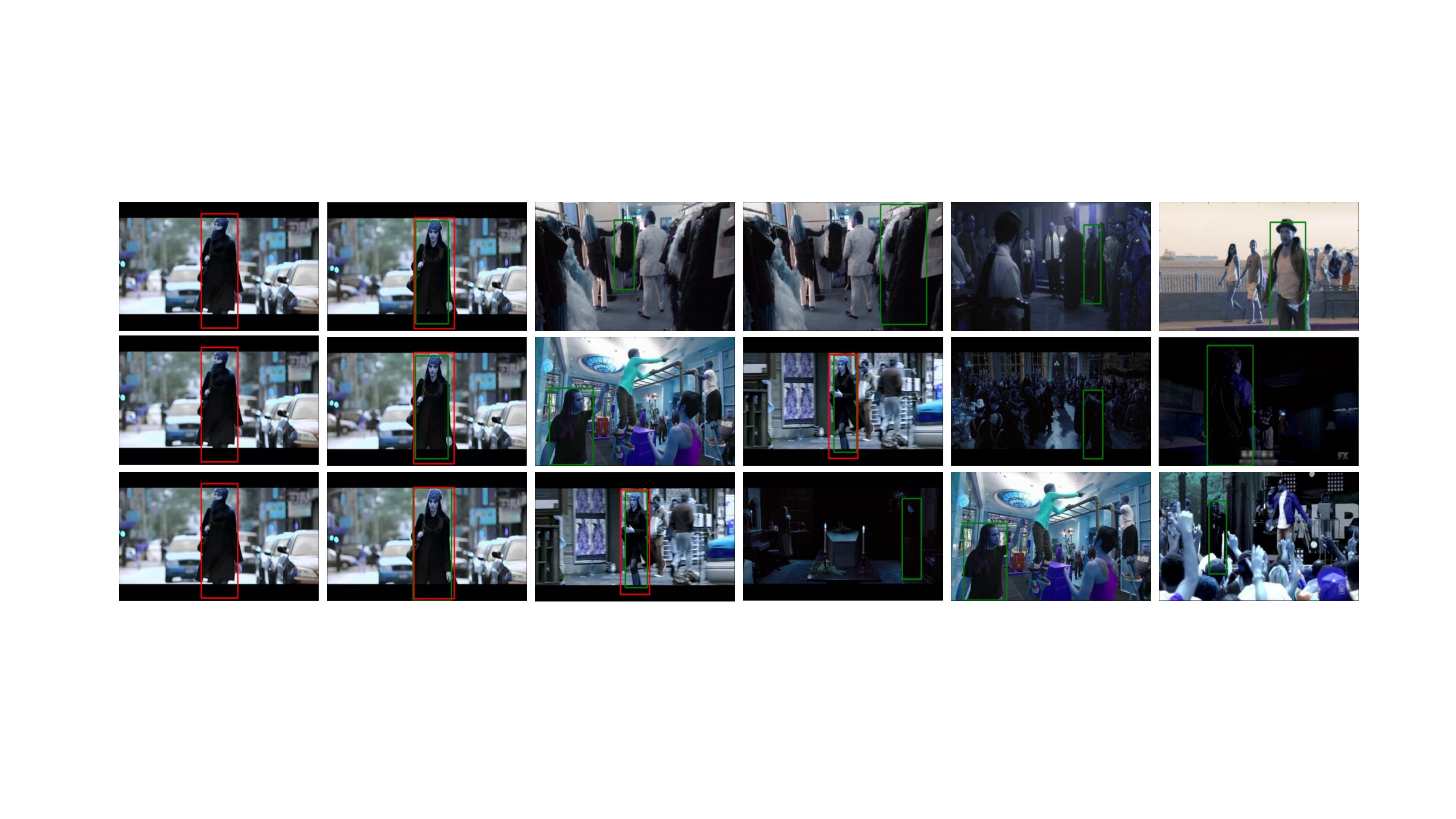} \\
\end{center}
   \caption{Three set of examples for the top-5 person search matches on the CUHK-SYSU test data, rows 1, 4, 7 are
   results of the E2E-PS \cite{xiao2016end}, rows 2, 5, 8 are
   results of E2E-PS \cite{xiao2016end} when ResNet-101 is used;  rows 3, 6, 9 are
   results of IAN. The red box region in the first column is the probe image. The green boxes
   in other columns are searching results, where red boxes are ground truth results. Best viewed in color.
    }
\label{fig:examples}
\end{figure*}

\vspace{3ex}
\noindent\textbf{Occlusion and Resolution.}
We also test IAN using low-resolution query persons
and partially occluded persons.
The gallery size is fixed as $100$ and several methods are evaluated
on these subsets. The results are shown in Table \ref{tab:resolution_occlusion}. It
is observed that all the methods perform significantly worse
on both the occlusion and low-resolution subsets than on the whole test set.
Nevertheless, IAN consistently outperforms E2E-PS \cite{xiao2016end} significantly.

\begin{table}
\begin{center}
\caption{Experimental results of three solutions on the occlusion
subset, low-resolution subset. }
\label{tab:resolution_occlusion}
\begin{tabular}{|c|c|c|c|c|c|c|}
\hline
\multirow{3}{*}{Method} & \multicolumn{2}{c|}{E2E-PS \cite{xiao2016end}} &   \multicolumn{2}{c|}{E2E-PS \cite{xiao2016end}} &  \multicolumn{2}{c|}{IAN}  \\
& \multicolumn{2}{c|}{VGGNet} & \multicolumn{2}{c|}{(Res-101)} & \multicolumn{2}{c|}{(Res-101)} \\
\cline{2-7}
& mAP & top-1 & mAP & top-1  & mAP & top-1 \\
\hline\hline
Low-Res & $46.11$ & $51.03$ &$47.91$ & $52.07$ & \textbf{52.60} & \textbf{54.48} \\
Occlusion & $44.33$ & $45.45$ &$47.79$  & $48.13$ & \textbf{53.02} & \textbf{54.55}  \\
Whole &  $69.69$ & $72.97$ & $74.28$  & $78.17$ & \textbf{77.23} & \textbf{80.45}  \\
\hline
\end{tabular}
\end{center}
\end{table}

\subsection{Results on PRW dataset}
\label{sec:prw}

The obtained results on the PRW dataset are reported in Table \ref{tab:prw_performance}.
Our proposed method outperforms the DPM-Alex+IDE$_{det}$  method reported in \cite{zheng2016person} with
a margin of more than $14\%$ top-1 accuracy.
More importantly, according to \cite{zheng2016person}, various ways of combining  of
pedestrian detection methods and re-identification methods are tested for the PRW dataset,
and it is shown that DPM-Alex+IDE$_{det}$ achieves the best performance among all the combinations.
On the other hand, the performance of IAN is also better than that of E2E-PS \cite{xiao2016end} and DPM-Alex+IDE$_{det}$,
which demonstrates the benefits of the center loss.

\begin{table}
\begin{center}
\caption{ Performance comparison on the PRW dataset with the state-of-the-art.}
\label{tab:prw_performance}
\begin{tabular}{|c|c|c|c|}
\hline
Method &  DPM-Alex  &  E2E-PS \cite{xiao2016end}  & IAN  \\
&  +IDE$_{det}$ \cite{zheng2016person}  & (ResNet-101) & (ResNet-101) \\
\hline\hline
mAP (\%)     &　$20.20$　&　$22.39$ &  \textbf{23.00} \\
top-1 (\%)   &　$48.20$　&  $61.00$ &  \textbf{61.85} \\
\hline
\end{tabular}
\end{center}
\end{table}

\section{Conclusions}
\label{sec:conclusion}

To address challenging issues in modern person search framework, we
proposed a novel Individual Aggregation Network (IAN) model that
can accurately localize pedestrians and meanwhile minimize intra-person variations
over feature representations. In particular, we built the IAN
upon the state-of-the-art object detection framework, i.e., faster R-CNN model,
so that high-quality region proposals for
pedestrians are produced in an online manner for person
search. In addition, IAN incorporates a novel center loss which is demonstrated
to be effective at relieving the negative effect caused by large variance of visual appearance of the same person.
Meanwhile, we also performed neural network compatibility
study for center loss, and we explained why dropout is not compatible with center loss.
Finally, extensive experiments on two benchmarks, i.e., CUHK-SYSU and PRW,
show that IAN achieves the state-of-the-art performance on both dataset, and
well demonstrate the superiority of the proposed IAN network.

One limitation of the proposed IAN is its large GPU memory requirement,
because center loss needs to track the feature centers of all classes.
Saving the GPU memory cost and reducing the network computational complexity will be our
future research work for the proposed IAN network.

\bibliographystyle{IEEEbib}

\end{document}